\documentclass[letterpaper,twocolumn,fleqn]{article} 

\usepackage{ist}

\usepackage{multirow}    %
\usepackage{booktabs}    %
\usepackage{array}       %
\usepackage{graphicx}    %
\usepackage{adjustbox}   %
\usepackage{caption}     %
\usepackage{xcolor}
\usepackage{amsmath}
\usepackage{amsfonts} 
\usepackage[numbers]{natbib}
\usepackage{hyperref}

\title{HistoWAS: A Pathomics Framework for Large-Scale Feature-Wide Association Studies of Tissue Topology and Patient Outcomes}

\author{ 
Yuechen Yang, Department of Computer Science, Vanderbilt University, Nashville, TN \\
Junlin Guo, Department of Computer Science, Vanderbilt University, Nashville, TN \\
Yanfan Zhu, Department of Computer Science, Vanderbilt University, Nashville, TN \\
Jialin Yue, Department of Radiation Oncology, Washington University in St. Louis, St. Louis, MO \\
Junchao Zhu, Department of Computer Science, Vanderbilt University, Nashville, TN \\
Yu Wang, Department of Biostatistics, Vanderbilt University Medical Center, Nashville, TN \\
Shilin Zhao, Department of Biostatistics, Vanderbilt University Medical Center, Nashville, TN \\
Haichun Yang, Department of Pathology, Microbiology and Immunology, Vanderbilt University Medical Center, Nashville, TN \\
Xingyi Guo, Department of Medicine, Vanderbilt University Medical Center, Nashville, TN \\
Jovan Tanevski, Institute for Computational Biomedicine, Heidelberg University and Heidelberg University Hospital, Heidelberg, Germany \\
Laura Barisoni, Department of Pathology, Duke University, Durham, NC \\
Avi Z. Rosenberg, Department of Pathology, Johns Hopkins University School of Medicine, Baltimore, MD\\
Yuankai Huo, Department of Computer Science, Vanderbilt University, Nashville, TN}

\date{}

\begin{document} 

\maketitle 

\thispagestyle{empty} 

\begin{figure*}[h]
    \centering
    \includegraphics[width=0.9\textwidth]{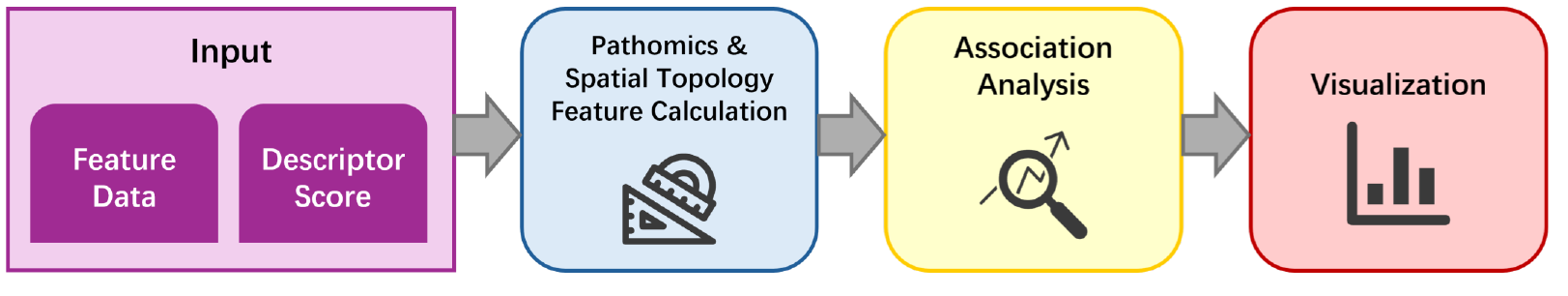} 
    \caption{An overview of the computational workflow for the HistoWAS framework. The process begins with two primary inputs: feature data extracted from WSIs and pathologist assigned histopathological descriptor scores. The framework then calculates a set of conventional object-level and spatial topology features. A high-throughput association analysis is subsequently performed to test the correlations between each feature and the clinical outcome. Finally, the results are interpreted and displayed using a suite of specialized visualizations.}
    \label{fig:pipeline}
\end{figure*}


\begin{abstract}
High-throughput "pathomic" analysis of Whole Slide Images (WSIs) offers new opportunities to study tissue characteristics and for biomarker discovery. However, the clinical relevance of the tissue characteristics at the micro- and macro-environment level is limited by the lack of tools that facilitate the measurement of the spatial interaction of individual structure characteristics and their association with clinical parameters. To address these challenges, we introduce HistoWAS (Histology-Wide Association Study), a computational framework designed to link tissue spatial organization to clinical outcomes. Specifically, HistoWAS implements (1) a feature space that augments conventional metrics with 30 topological and spatial features, adapted from Geographic Information Systems (GIS) point pattern analysis, to quantify tissue micro-architecture; and (2) an association study engine, inspired by Phenome-Wide Association Studies (PheWAS), that performs mass univariate regression for each feature with statistical correction. As a proof of concept, we applied HistoWAS to analyze a total of 102 features (72 conventional object-level features and our 30 spatial features) using 385 PAS-stained WSIs from 206 participants in the Kidney Precision Medicine Project (KPMP). The code and data have been released to \url{https://github.com/hrlblab/histoWAS}.

\end{abstract}

\begin{figure*}[t]
    \centering
    \includegraphics[width=0.9\textwidth]{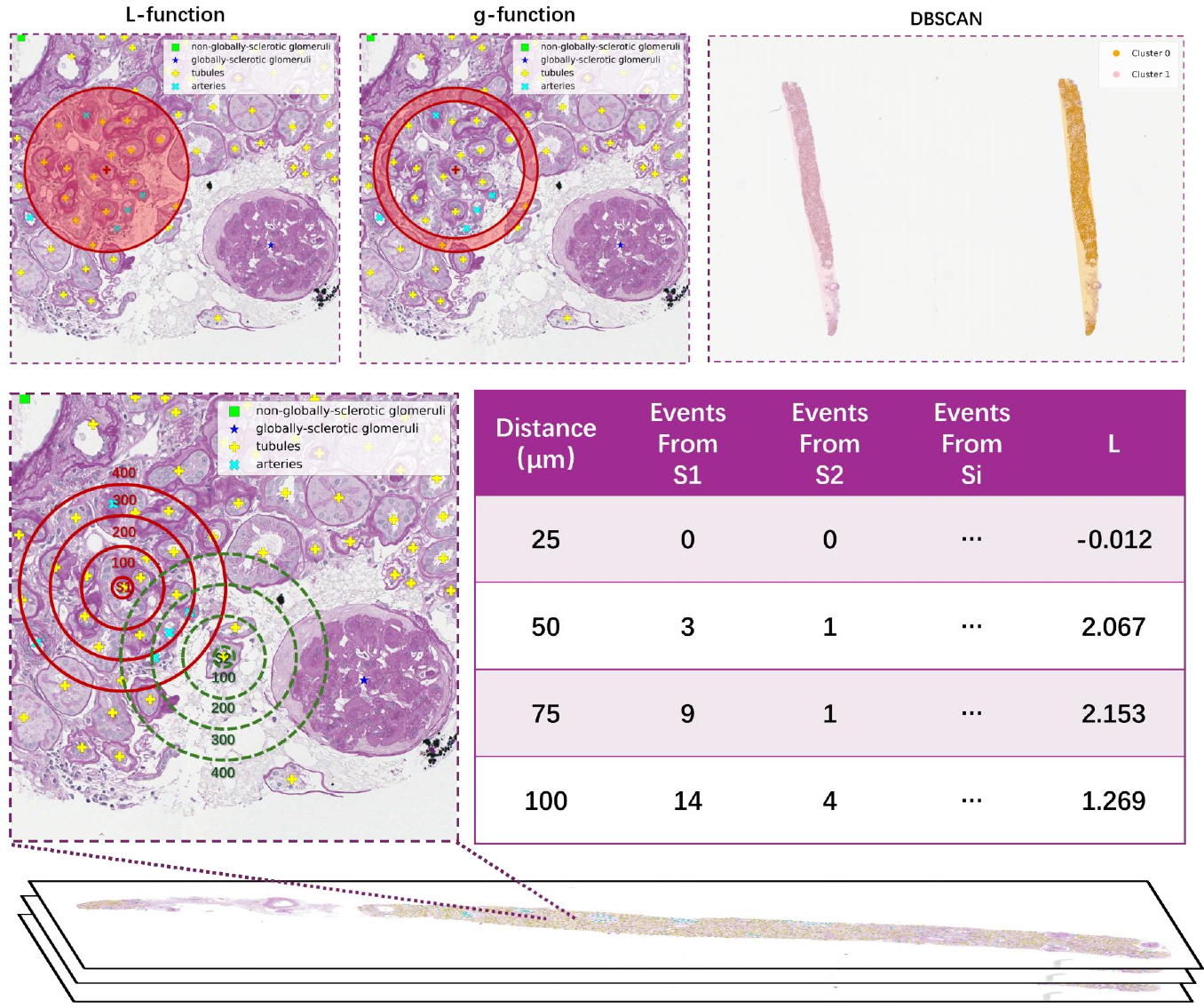}
    \caption{Conceptual workflow for spatial correlation feature extraction. (Top Right) The DBSCAN algorithm is used to identify and isolate distinct tissue clusters from the raw object point cloud, defining the region of interest for analysis. (Top Left \& Middle) Visual comparison between the cumulative L-function (counting all points within a radius r) and the non-cumulative Pair Correlation g-function (counting points in an annulus at a specific distance r). (Bottom) Demonstration of the distance-based function calculation. From a set of source points (S1,S2...Si), events (e.g., tubule centroids) are counted within expanding radii (e.g., 25, 50, 75 $\mu$m). These counts are then aggregated and normalized to compute a final feature value (L) for each distance, quantifying the tissue's micro-architectural pattern.}
    \label{fig:feature_extraction}
\end{figure*}


\section{1. Introduction}
\label{sec:intro}
The advent of digital pathology, centered around the analysis of WSIs, has driven a transition from qualitative, subjective assessment to quantitative, data-driven science\cite{lu2021integrating}. This transition has unlocked the field of "pathomics," the high-throughput extraction and analysis of large arrays of features from tissue morphology\citep{yang2025pyspatial,han2024development}.

While measuring characteristics of individual functional tissue units can enhance our understanding of individual biological processes\cite{stirling2021cellprofiler}, it is through a comprehensive analysis of the spatial interaction of these functional tissue units, at the tissue micro- and macroenvironment level, that ultimately will provide a more comprehensive, integrated, and perhaps physiologic, understanding of the kidney tissue ecosystem in healthy and diseased states. These spatial features are critical determinants of tissue function and pathological state. However, the development and deployment of tools capable of capturing the spatial complexity of tissue characteristics at the individual functional tissue unit, micro- and macroenvironment level and using this information for hypothesis generation and clinically actionable patient characterization, is still in its infancy.


Herein, we introduce HistoWAS, a computational framework designed to systematically identify digital biomarkers from the tissue micro-environment. The HistoWAS framework operates in two primary stages: (1) the extraction of a diverse pathomic feature set, including the aforementioned topological metrics, and (2) the execution of an association engine that tests for feature-phenotype associations with statistical correction. In this study, we used the Kidney Precision Medicine Project (KPMP) cases \citep{kpmp} and applied the HistoWAS framework to investigate the pathomic basis of renal interstitial fibrosis. Our analysis identifies several spatial features that are significantly associated with renal interstitial fibrosis, demonstrating the utility of the HistoWAS framework for systematic association analysis in digital pathology.

\section{2. Methods}

The HistoWAS framework employs a multi-stage computational pipeline, illustrated in Figure \ref{fig:pipeline}. Our objective was to develop a tool to link the spatial organization of tissues to clinical outcomes. We propose a paradigm integrating concepts from two distinct scientific domains. First, principles from GIS and spatial statistics are adapted to create a dictionary of interpretable topological features, such as metrics describing spacing, and inter-object correlation\cite{wu2023temporal}. These metrics extend beyond simple counts to quantitatively describe spatial patterns like aggregation, dispersion, and co-localization. Second, we adopt the high-throughput screening methodology of Genome-Wide (GWAS)\cite{uffelmann2021genome} and PheWAS Association Studies \cite{bastarache2022phenome, kerley2022pyphewas,kerley2023pyphewas} to perform a mass univariate analysis to test the association of each feature with a phenotype of interest. This section details the methods for topological feature extraction, the statistical approach for the association study, and the visualization techniques used for result interpretation.

\subsection{2.1 Spatial Feature Extraction}

A core component of the HistoWAS framework is the quantification of tissue micro-architecture. This was achieved by calculating a set of spatial and topological pathomic features for each WSI. As a conceptual example, Figure \ref{fig:feature_extraction} illustrates the calculation of our spatial correlation features.

\subsubsection{Tissue Area Definition}
Accurate estimation of the tissue area is critical for density-dependent spatial statistics. We developed an automated pipeline to calculate the total foreground tissue area for each WSI. First, all object centroids from a WSI were treated as a point cloud. The Density-Based Spatial Clustering of Applications with Noise (DBSCAN) algorithm was employed to identify tissue clusters and to separate them from slide artifacts and noise. For each resulting cluster containing at least three points, a convex hull was computed. The total tissue area for the WSI was then estimated by summing the areas of all such convex hulls (illustrated by the pink and orange regions in the top-right panel of Figure \ref{fig:feature_extraction}). This approach provides an approximation of the region of interest for subsequent spatial analyses.

\subsubsection{Spatial Point Pattern Features}

To characterize the spatial arrangement of histological objects, we computed a set of features derived from spatial point pattern analysis. This approach quantifies the tissue micro-architecture by modeling object centroids as spatial point processes.

Our framework extracts 30 distinct spatial pathomic features organized into three primary categories: (1) Density features (e.g., global density), (2) Spacing features (e.g., Average Nearest Neighbor), and (3) Correlation features (e.g., Besag's L-functions)\cite{liu2001comparison}.

A detailed technical description, including the mathematical formulation for each of these features, is provided in \textbf{\textit{Appendix}} section. All calculations included an edge correction method to account for boundary effects.

\subsection{2.2 Association Analysis}
\label{sec:association_analysis}
After feature extraction, we constructed an integrated feature matrix in which each row corresponds to an observation (e.g., a patient or WSI) and each column corresponds to a calculated feature. This matrix was then subjected to a high-throughput association study.

\subsubsection{Data Standardization}

Prior to statistical modeling, all feature values in the integrated matrix underwent a z-score transformation. This standardization step converted all features to a common scale (mean of 0, standard deviation of 1), to render comparable regression coefficients (beta values) across features with different native units and dynamic ranges.

\subsubsection{Statistical Model}

Modeled after the methodology of PheWAS, we performed a mass univariate linear regression analysis. For each feature, a separate linear model is fitted to test its association with the phenotype of interest. The model was specified as:
\begin{equation}
\text{Phenotype} \sim \beta_0 + \beta_1 \times \text{Feature}_i
\end{equation}
where $Phenotype$ represents the clinical outcome of interest and $\text{Feature}_i$ is the standardized value of the i-th feature. This process was repeated for all features. To account for multiple hypothesis testing, we calculated significance thresholds based on the Bonferroni correction ($\alpha = 0.05 / \text{number of tests}$) and the False Discovery Rate (FDR), using the Benjamini-Hochberg procedure with a Q-value of 0.05.

\subsection{2.3 Visualization}

To facilitate the output from the association study, we generated two visualizations, which are detailed in the \nameref{sec:visual} section:
\begin{itemize}
    \item \textbf{Manhattan Plot:} This plot was used to display the association results. Each feature is represented by a point, with its corresponding -log10(p-value) on the y-axis. To enhance visual clarity, the plot displays all features with -log10(p) values above the significance threshold, along with the top 25 most significant features falling below it. The x-axis plots the features, sorted by significance.
    
    \item \textbf{Effect Size Plot:} For features that passed the significance threshold, we created an effect size plot. This plot visualizes the beta coefficient and its 95\% confidence interval for each significant feature, providing a view of the magnitude and direction of the associations.
\end{itemize}

\begin{figure}[t]
    \centering
    \includegraphics[width=0.9\columnwidth]{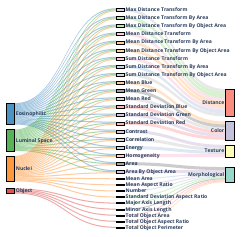}
    \caption{Organization of the 72 conventional object-level features provided by the KPMP. This Sankey diagram illustrates how quantitative metrics are derived from histological objects (left), such as Nuclei and Eosinophilic material, and grouped into four analytical categories: Morphological, Texture, Color, and Distance (right).}
    \label{fig:pathomics_data}
\end{figure}

\begin{figure}[t]
    \centering
    \includegraphics[width=0.9\columnwidth]{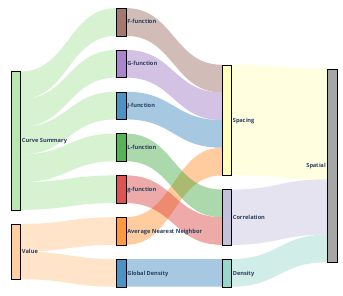}
    \caption{Hierarchical organization of the 30 spatial features. This Sankey diagram illustrates the categorization of the spatial metrics used in the HistoWAS framework. It shows how specific feature extraction methods (left) are grouped into analytical categories (middle) and combined into the final "Spatial" feature set (right).}
    \label{fig:spatial_data}
\end{figure}

\begin{figure*}
    \centering
    \includegraphics[width=0.9\textwidth]{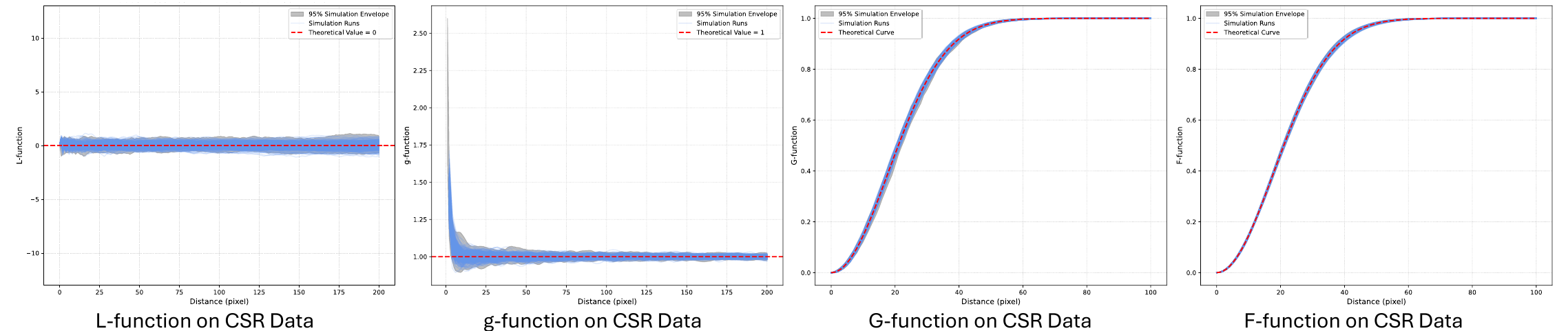}
    \caption{Validation of the spatial feature extraction pipeline using simulated 
    CSR data. (Left to Right) The plots for the L-function, g-function, G-function, and F-function are shown. In all four cases, the results from hundreds of individual simulation runs (blue lines) fall within the 95\% simulation envelope (gray shaded area) and align with the theoretical values (red dashed line). The theoretical values are $L(r)=0$, $g(r)=1$, and $G(r) = F(r) = 1 - \exp(-\lambda \pi r^2)$, respectively.}
    \label{fig:robutness}
\end{figure*}

\begin{figure}
    \centering
    \includegraphics[width=0.9\columnwidth]{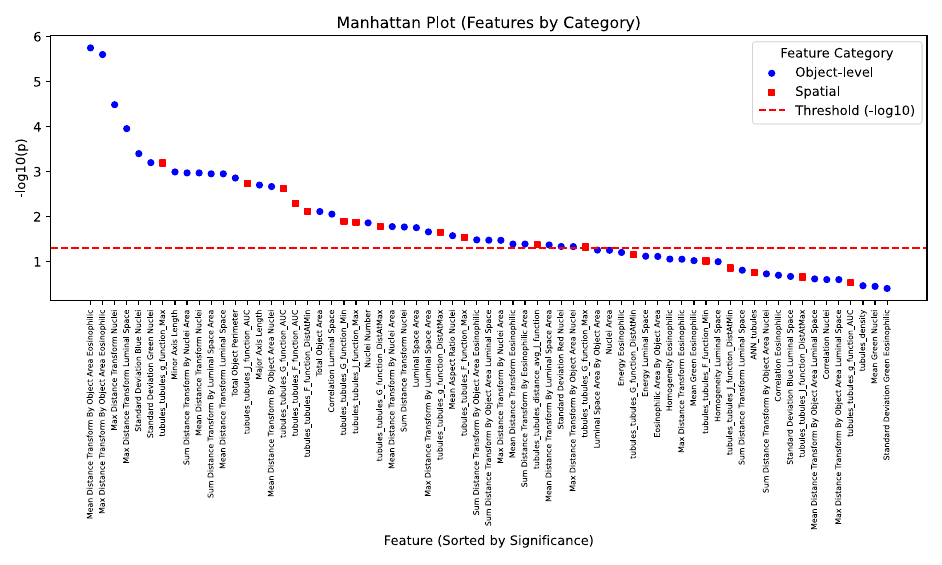}
    \caption{Manhattan plot of HistoWAS association results. This plot displays the statistical significance for all features that passed the significance threshold ($p<0.05$), as well as the top 25 features that fell below this threshold. All features shown are sorted by their p-value along the x-axis. The y-axis represents the $-\log_{10}$(p-value), meaning that higher points on the plot correspond to stronger statistical significance. The horizontal dashed line indicates the predefined significance threshold; any features extending above this line are considered statistically significant. Features are color-coded by their category (Object-level in blue, Spatial in red) to distinguish between the feature types.}
    \label{fig:Manhattan}
\end{figure}

\section{3. Experiments}

\subsection{3.1 Datasets}
The dataset utilized in this study was sourced from the KPMP cohort\cite{kpmp}, comprising a total of 385 WSIs from 206 patients. These periodic acid Schiff-stained WSIs were acquired or processed in five distinct batches. The number of scans in each batch was as follows: Batch 1 contained 236 scans, Batch 2 contained 62 scans, Batch 3 contained 26 scans, Batch 4 contained 42 scans, and Batch 5 contained 19 scans.

The clinical descriptor scores were generated by the KPMP Pathology Descriptor Scoring Task Force using a standardized, multi-stage protocol. This process was initiated by a primary scorer who performed an assessment using images from the KPMP Digital Pathology Repository. Following this, a dedicated Quality Control (QC) scorer independently reviewed all scores. When discrepancies occurred, scores were adjudicated between the primary and QC scorers to ensure consensus and data reliability. The resulting dataset comprises 153 distinct descriptor columns, with data stratified by anatomical sections, including Cortex, Medulla, and combined Cortex and Medulla.

We selected Interstitial fibrosis visually assessed by KPMP study pathologists, as primary endpoint (Descriptor Score) for the Histo-Wide Association Study. A single continuous (\%) fibrosis score was available for each of the 206 patients, who constituted the final cohort for our analysis.

To align the WSI-level features with the patient-level clinical endpoint, a two-stage feature extraction process was employed. First, a total of 102 quantitative features were extracted from each of the 385 WSIs. This feature set was categorized into two groups: 72 object-level features (representing morphological and statistical metrics)  and 30 spatial features (describing topological organization). Second, for patients with multiple WSIs, these WSI-level feature vectors were aggregated (e.g., by calculating the mean) to create a single feature vector for each patient. This resulted in a final analysis matrix of 206 patients by 102 features.

\subsection{3.2 Object-level features}
This study utilized 72 morphological object-level features extracted from the KPMP WSIs. As illustrated in Figure \ref{fig:pathomics_data}, these pre-computed features quantitatively characterize histological objects derived from tissue components including nuclei, luminal space, and eosinophilic material. The features are categorized into four aspects: (1)Morphological features, such as Major/Minor Axis Length and Mean Aspect Ratio, quantify the geometric and shape properties of the detected objects. (2) Haralick texture features, including Correlation, Energy, and Homogeneity, were included to describe pixel-level uniformity and complexity. (3) Color-based features were extracted to quantify staining properties, capturing first-order statistics like the Mean and Standard Deviation for each Red, Green, and Blue color channel. (4) Features based on the distance transform, such as Mean Distance Transform and Max Distance Transform, were calculated to describe the spatial relationships of pixels within an object, providing metrics for object compactness and thickness.

\subsection{3.3 Spatial features}
To quantitatively capture the micro-architectural organization of tissues, the HistoWAS framework extracts 30 spatial features. These features are designed to characterize the collective arrangement and interrelationships of the object population.

As visualized in the Sankey diagram (Figure \ref{fig:spatial_data}), these 30 features are hierarchically organized. They are synthesized from 8 spatial analysis methods, which are grouped into three analytical categories: Correlation, Spacing, and Density.

These categories summarize the tissue's spatial properties. For instance, Correlation features quantify inter-object relationships by leveraging second-order point pattern analysis (e.g., K, L, and g-functions), which are used to identify non-random patterns such as clustering or repulsion. Spacing features provide a set of features that quantify the point distribution by summarizing nearest-neighbor distances (e.g., G-function) and empty-space distances (e.g., F-function). Finally, Density features provide a measure of object concentration (e.g., Global Density).

The set of 30 features is generated by applying these methods in a multi-scale fashion. As indicated by the "Curve Summary" inputs in the diagram, many of these metrics are not single scalar values. Instead, they represent a summary statistic (e.g., mean, std) from their respective feature.

\subsection{3.4 Simulation}
We conducted a simulation study to test our implementation of the feature extraction pipeline. The goal was to determine if our implementation computes key spatial functions in accordance with known theoretical values under conditions of CSR.

We first generated a large-scale synthetic point cloud following a homogeneous CSR (Complete Spatial Randomness) process (i.e., a Poisson point process) with a known global density, $\lambda$. From this large-scale "base" image, we simulated an analysis scenario by repeatedly sampling hundreds of smaller sub-regions (N=299 samples), analogous to individual WSIs or tissue patches. 

For each of these 299 sampled point patterns, we executed our HistoWAS pipeline to calculate the four distance-based functions: the L-function, g-function, G-function, and F-function, applying the edge corrections described in the \nameref{sec:edge_correction} section.

Under CSR, these functions have well-defined theoretical values. The centered L-function, $L(r)$, should be 0 for all distances $r$. The Pair Correlation g-function, $g(r)$, should be 1. The nearest-neighbor G-function, $G(r)$, and the empty-space F-function, $F(r)$, should both follow the theoretical cumulative distribution curve $F(r) = G(r) = 1 - \exp(-\lambda \pi r^2)$.

\subsection{3.5 Association Analysis on Empirical Data}

We applied the HistoWAS association analysis framework, as described in \nameref{sec:association_analysis} section, to the KPMP cohort data.

The interstitial fibrosis score was used as the single Phenotype (dependent variable) in our statistical model. This score was available at the patient level ($N=206$).

As an example, the independent variables Feature$_i$ in the model comprised the complete set of 102 standardized features (72 object-level features and 30 spatial features). One step in the experiment was aligning the WSI-level features with the patient-level clinical endpoint. The 102 quantitative features were first extracted from each of the 385 WSIs. Then, for patients with multiple WSIs, these feature vectors were aggregated (e.g., by calculating the mean) to create a single, unified feature vector for each of the 206 patients. This resulted in a final analysis matrix of 206 patients by 102 features.

Following data standardization, each of the 102 features was then tested for association against the interstitial fibrosis score using the defined univariate regression model.

\begin{figure}[t]
    \centering
    \includegraphics[width=0.8\columnwidth]{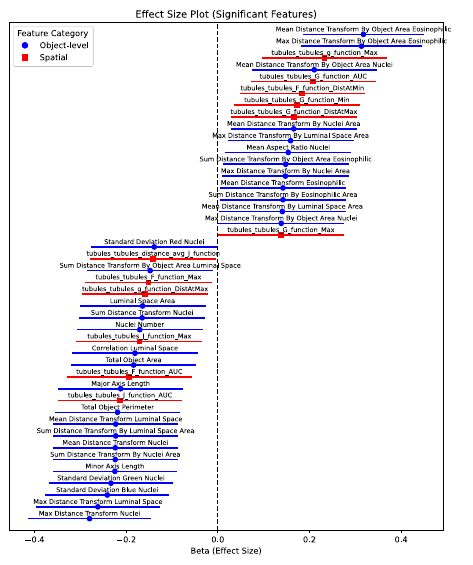}
    \caption{Effect size plot of significant features identified by HistoWAS. This plot provides a summary of the features that passed the significance threshold. Each feature is listed on the y-axis, with its corresponding effect size (Beta coefficient) and 95\% confidence interval plotted on the x-axis. Features on the right ($Beta > 0$) are positively associated with the clinical outcome, while those on the left ($Beta < 0$) are negatively associated. This visualization allows for a comparison of the magnitude, direction, and statistical precision of each significant association. Features are color-coded by category (Pathomics in blue, Spatial in red).}
    \label{fig:Effectsize}
\end{figure}

\section{4. Results}
\subsection{4.1 Simulation Results}
The results of this simulation are presented in Figure~\ref{fig:robutness}. The plots show the calculated curves from all 299 simulation runs (blue lines) overlaid against the theoretical 95\% simulation envelope (gray area) and the known theoretical values (red dashed line). 

As shown in the results, the calculated L-function, g-function, G-function, and F-function values remained overwhelmingly within the expected confidence intervals and aligned closely with their theoretical ground truths. These findings indicate that our feature extraction pipeline computes the underlying spatial statistics in accordance with their theoretical definitions.

\subsection{4.2 Association Analysis}
\label{sec:visual}
The HistoWAS association results are presented in two visualizations.

A Manhattan plot (Figure \ref{fig:Manhattan}) provides an overview of the statistical significance for all tested features. In this plot, the y-axis represents the $-\log_{10}(p\text{-value})$, where a higher point on the plot corresponds to a stronger statistical significance (i.e., a smaller $p$-value), allowing for the identification of feature associations that surpass the predefined significance threshold.

An effect size plot (Figure \ref{fig:Effectsize}) displays the effect size (beta coefficient) and 95\% confidence interval for each significant feature, illustrating the direction, magnitude, and precision of its association with the clinical outcome. In this plot, the direction is shown by whether the coefficient is positive (on the right of the zero line) or negative (on the left). The magnitude (or strength) of the association is represented by the coefficient's distance from zero. Finally, the precision is indicated by the width of the 95\% confidence interval (the horizontal bar), where narrower bars signify a more precise estimate of the effect.

\section{5. Discussion}

A primary contribution of HistoWAS is its computational framework for linking interpretable spatial features to clinical outcomes. By design, the framework uses spatial statistics (such as patterns of cellular clustering or repulsion at specific distances). This approach identifies specific, quantifiable spatial patterns associated with the outcome, which can serve as a basis for generating testable biological hypotheses.

There are some limitations that should be acknowledged to guide future research. First,  this is a pilot study, as a proof of concept, using Interstitial Fibrosis visual score as the sole histological phenotype. Furthermore, the scope of our analysis was constrained by only selecting tubules for feature computation. Future efforts could broaden the range of visual scores and tissue structures for a more comprehensive spatial analysis, and clinical outcomes such as estimated Glomerular Filtration Rate (eGFR) or Albumin-to-Creatinine Ratio (ACR). Second, we used ROIs that were approximated computationally using DBSCAN and convex hulls. Future implementations with WSIs will allow for more accurate ROI delineation. Future research should integrate biological experiments to elucidate the mechanisms and clinical relevance of these spatial patterns, thereby maximizing the translational potential of HistoWAS.


\section{6. Conclusion}

HistoWAS is a computational framework designed to associate interpretable spatial features from the tissue micro-architecture of digital pathology images with clinical outcomes. By integrating object-level morphometric pathomic features with multi-scale spatial topology features, and performing a high-throughput association study against visual scores of tissue damage, our framework identified multiple spatially-defined patterns that are associated with the disease state.

\section{Acknowledgment}
This research was supported by NIH R01DK135597 (Huo), DoD HT9425-23-1-0003 (HCY), NSF 2434229 (Huo) and KPMP Glue Grant. This work was also supported by Vanderbilt Seed Success Grant and Vanderbilt-Liverpool Seed Grant. This research was also supported by NIH grants R01EB033385, R01DK132338. We extend gratitude to NVIDIA for their support by means of the NVIDIA hardware grant. 

The Kidney Precision Medicine Project (KPMP) is supported by the National Institute of Diabetes and Digestive and Kidney Diseases (NIDDK) through the following grants: U01DK133081, U01DK133091, U01DK133092, U01DK133093, U01DK133095, U01DK133097, U01DK114866, U01DK114908, U01DK133090, U01DK133113, U01DK133766, U01DK133768, U01DK114907, U01DK114920, U01DK114923, U01DK114933, U24DK114886, UH3DK114926, UH3DK114861, UH3DK114915, and UH3DK114937. We gratefully acknowledge the essential contributions of our patient participants and the support of the American public through their tax dollars.


\small

\bibliographystyle{unsrt}

\bibliography{references}


\begin{biography}
Yuechen Yang is a second-year PhD student in Computer Science Department at Vanderbilt University. Her research focuses on medical image processing, with a particular emphasis on the extraction and application of pathomics.
\end{biography}

\clearpage
\section{Appendix: Spatial Feature Extraction}
\label{sec:appendix}

To quantitatively characterize the tissue micro-architecture, the HistoWAS framework extracts spatial features. As illustrated in Figure~\ref{fig:spatial_data}, these features are derived from the centroid coordinates of histological objects and are organized into three analytical categories: Density, Spacing, and Correlation.

\subsection{Density Features}
Density features describe the concentration of histological objects within the tissue area. This category includes Global Density, which is a single-value metric representing the total count of a specific object type (e.g., tubules) divided by the total estimated tissue area.


\subsection{Correlation Features}
Correlation features provide a set of features that analyze the spatial dependencies between points, quantifying patterns like clustering or dispersion at multiple distance scales. We then extract summary statistics from the resulting L-function and g-function curves (detailed in the \nameref{sec:curve_summary} section).

\subsubsection{Besag's L-function}

To quantify spatial patterns such as clustering, we use Besag's L-function, $L(r)$. The L-function is a variance-stabilized transformation of Ripley's K-function, $K(r)$, which is centered at 0 for a completely random pattern, making it easier to interpret\cite{baddeley2016spatial}.

The K-function is a cumulative metric that summarizes the expected number of points within a distance $r$ of any given point, normalized by the overall density $\lambda$.
\begin{equation}
    \hat{K}(r) = \frac{A}{n^2} \sum_{i=1}^{n} \sum_{j=1, j\neq i}^{n} w_{ij} \cdot \mathbb{I}(d_{ij} \le r)
\end{equation}
where $A$ is the total area, $n$ is the number of points, $d_{ij}$ is the distance between points $i$ and $j$, $\mathbb{I}(\cdot)$ is the indicator function, and $w_{ij}$ is an edge-correction weight. We then transform $K(r)$ into the variance-stabilized Besag's L-function, $L(r)$, which is centered at 0 for a random pattern.
\begin{equation}
    \hat{L}(r) = \sqrt{\frac{\hat{K}(r)}{\pi}} - r
\end{equation}
A value of $L(r) > 0$ at a given $r$ indicates clustering at that scale.

\subsubsection{g-function}
While the K-function is cumulative, the g-function, $g(r)$, is a non-cumulative density-based metric, as illustrated in Figure \ref{fig:feature_extraction}. It measures the probability of finding two points separated by a specific distance $r$, relative to the probability expected under CSR. It is formally related to the K-function:
\begin{equation}
    g(r) = \frac{1}{2\pi r} \frac{dK(r)}{dr}
\end{equation}
For a random pattern, $g(r) = 1$. A value of $g(r) > 1$ indicates aggregation at distance $r$.

\begin{figure}[t]
    \centering
    \includegraphics[width=0.9\columnwidth]{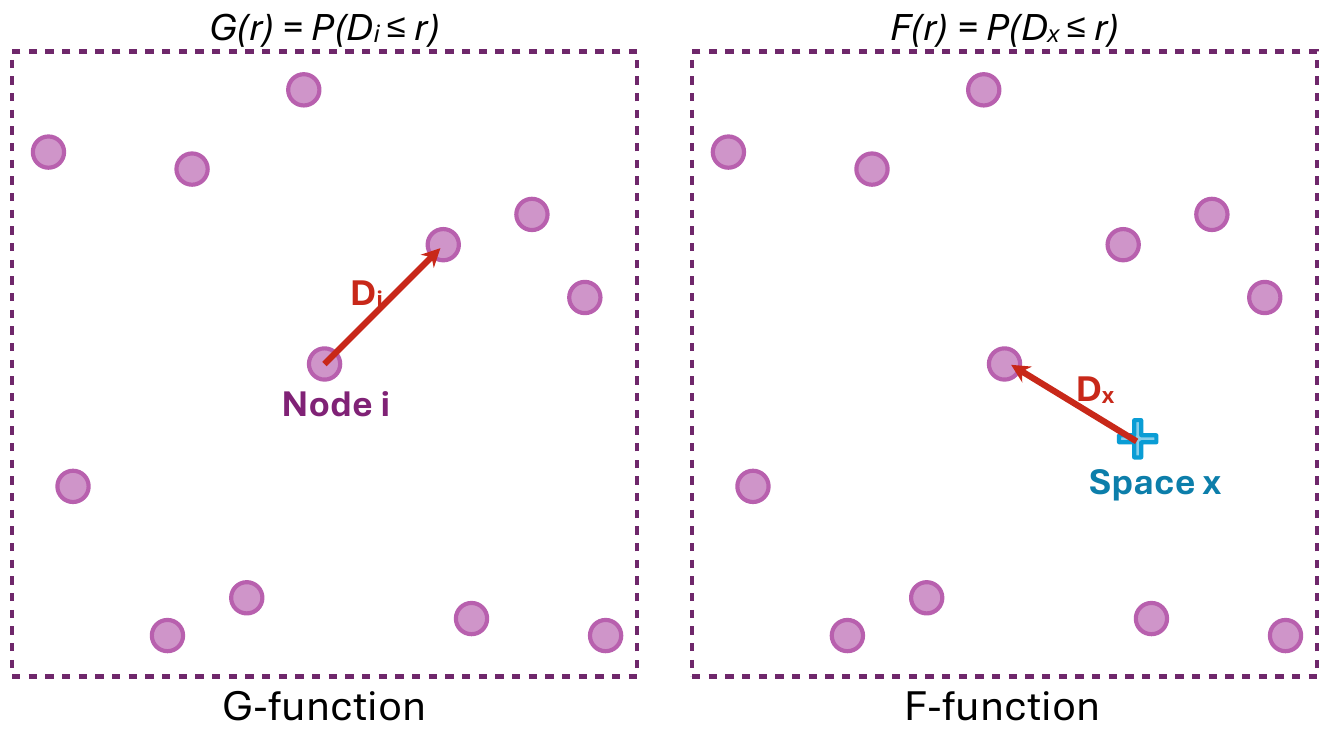}
    \caption{Conceptual comparison of the G-function and F-function\cite{baddeley2016spatial}. This diagram illustrates the difference between two spatial statistics. (Left) The G-function, or nearest-neighbor distance function, measures the distribution of distances from each point in the pattern (Node i) to its closest neighboring point. (Right) In contrast, the F-function, or empty space function, measures the distribution of distances from a random location in the study area (Space x) to the closest point in the pattern . This distinction allows the G-function to characterize point-to-point spacing, while the F-function characterizes the size of the empty spaces within the pattern\cite{baddeley2016spatial}.}
    \label{fig:GFfunction}
\end{figure}

\begin{figure}[t]
    \centering
    \includegraphics[width=0.9\columnwidth]{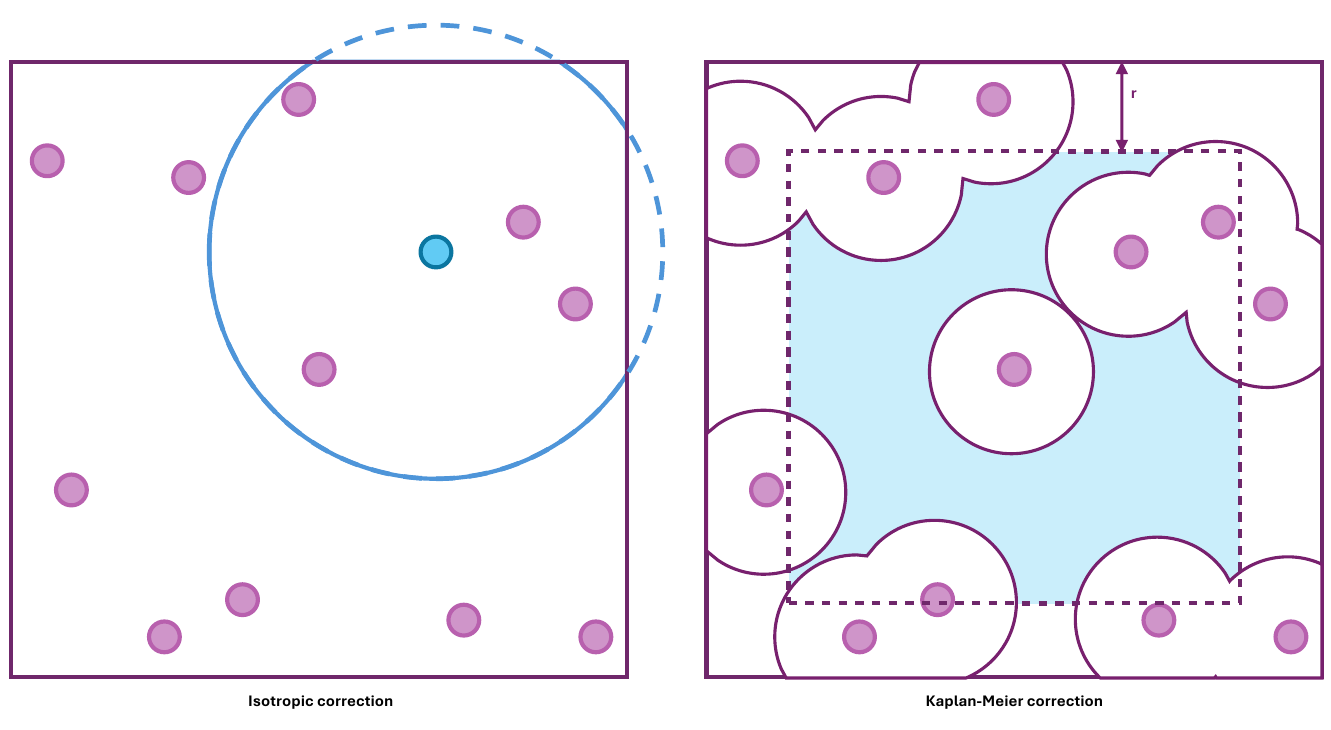}
    \caption{Conceptual illustration of edge effects and correction methods\cite{baddeley2016spatial}. 
    (Left) The "edge effect" problem: The search circle (solid blue) for a point (cyan) near the boundary extends into an unobserved area (outside the purple box). Simply counting points within the observed part of the circle leads to underestimation. Ripley's isotropic correction compensates for this by weighting the observation based on the proportion of the circle's circumference that lies inside the window.
    (Right) Censoring in nearest-neighbor functions (like the F-function): The Kaplan-Meier estimator handles "censored" observations, such as when the distance from a random location to the nearest point is greater than its distance to the boundary (dashed line). The true nearest neighbor might be in the unobserved region\cite{baddeley2016spatial}.}
    \label{fig:edge_correction}
\end{figure}

\subsection{Spacing Features}
Spacing features\cite{baddeley2016spatial} provide a set of features that quantify the point distribution by summarizing nearest-neighbor distances (e.g., G-function) and empty-space distances (e.g., F-function).

\subsubsection{G-function (Nearest-Neighbor Distance Function)}
The G-function, $G(r)$, is the cumulative distribution function (CDF) of the distances from each point to its nearest neighbor\cite{soltisz2024spatial}. It estimates the probability that a point's nearest neighbor is within distance $r$, which is illustrated in Figure \ref{fig:GFfunction} (left).
\begin{equation}
    G(r) = P(D_i \le r)
\end{equation}
where $D_i$ is the distance from point $i$ to its nearest neighbor.

\subsubsection{F-function (Empty Space Function)}
The F-function, $F(r)$, is the "empty space" or "point-to-nearest-event" function. It is the CDF of the distances from a set of randomly sampled locations (quadrat points) in the study area to the nearest data point, which is illustrated in Figure \ref{fig:GFfunction} (right).
\begin{equation}
    F(r) = P(Y_x \le r)
\end{equation}
where $Y_x$ is the distance from a random location $x$ to the nearest point in the pattern.

\subsubsection{J-function}
The J-function provides a combined summary of the G and F functions, defined as:
\begin{equation}
    J(r) = \frac{1 - G(r)}{1 - F(r)} \quad \text{for } F(r) < 1
\end{equation}
For a CSR pattern, $J(r) = 1$. $J(r) < 1$ indicates clustering, while $J(r) > 1$ indicates regularity or dispersion.

\subsubsection{Average Nearest Neighbor(ANN)} 
The ANN metric characterizes the global clustering or dispersion of a pattern by calculating the average distance from each point to its single closest neighbor\cite{cressie2015statistics}.

\begin{equation}
    d_{ANN} = \frac{\sum_{i=1}^{n} \min(d_{ij})}{n}
\end{equation}
where $n$ is the total number of points, and $\min(d_{ij})$ is the distance between point $i$ and its nearest neighboring point $j$.

\subsection{Curve Summary Features}
\label{sec:curve_summary}
For the distance-based functions, we derive features from the overall shape of the centered curves. For each function (L, g, G, F, J), we calculate a set of summary statistics, including:
\begin{itemize}
    \item \textbf{AUC:} The Area Under the Curve of the centered function, measuring the total cumulative deviation from randomness across all distances.
    \item \textbf{Max/Min:} The maximum positive (clustering) and minimum negative (dispersion) peaks of the centered curve.
    \item \textbf{DistAtMax/DistAtMin:} The specific distances ($r$) at which these maximum and minimum deviations occur, identifying the characteristic scales of clustering or dispersion.
\end{itemize}

\subsection{Edge Effect Correction}
\label{sec:edge_correction}

Calculating spatial statistics within a finite observation window (i.e., the estimated tissue area) introduces a systemic bias known as the edge effect. As illustrated in Figure~\ref{fig:edge_correction} (Left), a point located near the window's boundary may have its neighborhood (e.g., a circle of radius $r$) extend beyond the observed area. Failing to account for this unobserved portion leads to a biased underestimation of point counts or neighbor distances.

To account for this, the HistoWAS framework employs two function-specific correction methods:

\begin{itemize}
    \item \textbf{Ripley's Isotropic Correction (for $L$ and $g$-functions):} The L-function and g-function are derived from pair-counting statistics. To correct the underlying calculations for these functions (specifically, the K-function and g-function), we apply the isotropic correction\cite{ge2019geographic}. This method computes the proportion of the circumference of a circle of radius $r$ centered on a point $i$ that lies within the observation window. It then weights the contribution of point pairs at this distance by the inverse of this proportion, compensating for unobserved neighbors.

    \item \textbf{Kaplan-Meier Estimator (for $G$, $F$, and $J$-functions):} For nearest-neighbor distance functions, a non-parametric "survival" analysis is used. We use the Kaplan-Meier estimator\cite{diggle2013statistical}. This method treats observations as either "censored" or "uncensored". 
    For the G-function, an observation is "censored" if a point's distance to the boundary is smaller than its distance to its nearest observed neighbor, implying the true nearest neighbor might exist in the unobserved area.
    For the F-function, the same logic applies to random sample points (Figure~\ref{fig:edge_correction}, Right). The Kaplan-Meier method estimates the CDF of nearest-neighbor distances in the presence of this censoring. The J-function, being a ratio of $G$ and $F$, inherits this correction.
\end{itemize}

\end{document}